\documentclass[table, 11pt]{article}
\usepackage[table, svgnames, dvipsnames]{xcolor} % Gộp xcolor
\usepackage{tikz}
\usepackage[justification=centering]{caption}
\usepackage[most]{tcolorbox}
\colorlet{promptbg}{green!15}
\usepackage[colorlinks=true, citecolor=blue, linkcolor=blue, urlcolor=blue]{hyperref}
\usepackage[final]{acl}
\usepackage{amsfonts} % For \mathbb{R}
\usepackage{booktabs} % For better table rules
\usepackage{multirow} % If needed for table structure, though not strictly for this one
\usepackage{graphicx} % ở phần preamble nếu chưa có
\usepackage{amsfonts} % For \mathbb{R}
\usepackage{booktabs} % For professional tables
\usepackage{multirow} % For multi-row cells in tables
\usepackage{graphicx} % Required for includegraphics (though not used in text here)
\usepackage{xcolor} % For coloring text
\usepackage{bold-extra} % For bold text if needed, or just use \textbf
\usepackage{times}
\usepackage{latexsym}
\usepackage{enumitem}
\usepackage[T1]{fontenc}
\usepackage[utf8]{inputenc}
\usepackage{microtype}
\usepackage{inconsolata}
\usepackage{graphicx}
\usepackage{amsmath}
\usepackage{amsfonts}
\usepackage{bm}
\usepackage{multirow}
\usepackage{booktabs}
\usepackage{xspace}
\usepackage{adjustbox}
\usepackage{float}
\usepackage{stfloats}
\usepackage{natbib} % Thêm natbib để hỗ trợ \citep, \citet

\usepackage{colortbl}
\usepackage{subfigure}

\usepackage[linesnumbered,ruled,lined]{algorithm2e}

\definecolor{lightgray}{gray}{0.80}
\definecolor{darkgreen}{rgb}{0.0, 0.5, 0.0}
\newcolumntype{g}{>{\columncolor{lightgray}}c}

\title{SRA: Span Representation Alignment for Large Language Model Distillation}

\author{
  \textbf{Quoc Phong Dao\textsuperscript{1}\footnotemark[1]},
    \textbf{Hoang Son Nguyen\textsuperscript{1}\footnotemark[1]},
  \textbf{Pham Khanh Chi\textsuperscript{1}\footnotemark[1]},
   \textbf{Tung Nguyen\textsuperscript{1}}, \\
    \textbf{Linh Ngo Van\textsuperscript{1,\dag}},
  \textbf{Diep Thi-Ngoc Nguyen\textsuperscript{2}},
  \textbf{Trung Le\textsuperscript{3}}
  \bigskip \\
\textsuperscript{1}Hanoi University of Science and Technology, \\
\textsuperscript{2}VNU University of Engineering and Technology,
\textsuperscript{3}Monash University
}

\begin{document}

\maketitle
\renewcommand{\thefootnote}{\fnsymbol{footnote}}
\footnotetext[1]{Equal contribution}
\footnotetext[2]{Corresponding author: \href{mailto:email@domain}{ linhnv@soict.hust.edu.vn}}
\renewcommand*{\thefootnote}{\arabic{footnote}}
\begin{abstract}
    Cross-Tokenizer Knowledge Distillation (CTKD) enables knowledge transfer between a large language model and a smaller student, even when they employ different tokenizers. While existing approaches mainly focus on token-level alignment strategies, which are often brittle and sensitive to discrepancies between tokenizers, we argue that the method of aggregating tokens into more robust representations before distillation is of equal importance. In this paper, we introduce \textbf{SRA} (\textbf{S}pan \textbf{R}epresentation \textbf{A}lignment for Large Language Model Distillation), a novel framework that reframes CTKD through the physical lens of Multi-Particle Dynamical Systems. SRA shifts the fundamental unit of alignment from tokens to robust, tokenizer-agnostic spans. We model each span as a cluster of particles and represent its state by its Center of Mass (CoM) - an attention-weighted average that captures rich semantic information. We leverage the concept of span centers of mass with attention-derived weighting to prioritize the most salient spans. In addition, we employ a geometric regularizer to preserve the structural integrity of the representation space and introduce aligned span logit distillation to enhance knowledge transfer across models. In challenging cross-architecture distillation experiments, SRA consistently and significantly outperforms state-of-the-art CTKD baselines, validating our physically-grounded approach.
\end{abstract}
\section{Introduction}\label{sec:intro}
Large Language Models (LLMs) have achieved remarkable success, largely driven by scaling model parameters to billions or even trillions \citep{deepseekv32024, openaigpt42024}. However, this immense scale poses significant challenges for practical deployment. Knowledge Distillation (KD), which transfers knowledge from a large teacher to a smaller student, has emerged as a critical technique for creating efficient yet powerful models \citep{hinton2015distillingknowledgeneuralnetwork}. While effective, conventional KD often assumes the teacher and student share an identical tokenizer, a restrictive assumption in practice \citep{sun2019patientknowledgedistillationbert, sanh2020distilbertdistilledversionbert, gu2024minillmknowledgedistillationlarge}. Existing Cross-Tokenizer Knowledge Distillation (CTKD) methods tackle these mainly at the \emph{token or logit level}: unifying output spaces via projections or cross-model attention \citep{zhang2024dualspaceknowledgedistillationlarge},
transporting probability mass between vocabularies with Optimal Transport \citep{boizard2025crosstokenizerdistillationuniversallogit,cui2025multi},
 or aligning token streams by edit distance \citep{wan2024knowledgefusionlargelanguage,chen2025enhancingcrosstokenizerknowledgedistillation}.
 
 We argue that a more robust solution requires a paradigm shift, grounding CTKD in the underlying dynamics of the Transformer architecture itself. The Transformer, with its layer-wise skip connections, can be interpreted as a discretized Ordinary Differential Equation (ODE), where each layer functions as a discrete step in time \citep{chen2018neuralode, lu2019understanding}. In this ODE framework, the teacher’s deep architecture provides a fine-grained discretization of the feature dynamics, while the student’s shallower architecture is a coarser one. This perspective is powerfully complemented by the interpretation of Transformers as \textbf{Multi-Particle Dynamical Systems (MPDS)} \citep{lu2019understanding}, which models tokens as "particles" whose states (hidden representations) evolve through time (layers). Together, these views establish a clear principle: distillation should not be limited to final outputs, but must instead focus on transferring the \textit{dynamics of the system's geometric evolution}. Recent work has begun to explore this, moving beyond static representations to distill the feature dynamics themselves \citep{gong2025beyond}. However, such methods are confined to same-tokenizer settings and do not address the fundamental geometric and granularity challenges of CTKD.

\begin{figure}[t]
\centering
\includegraphics[width=\linewidth]{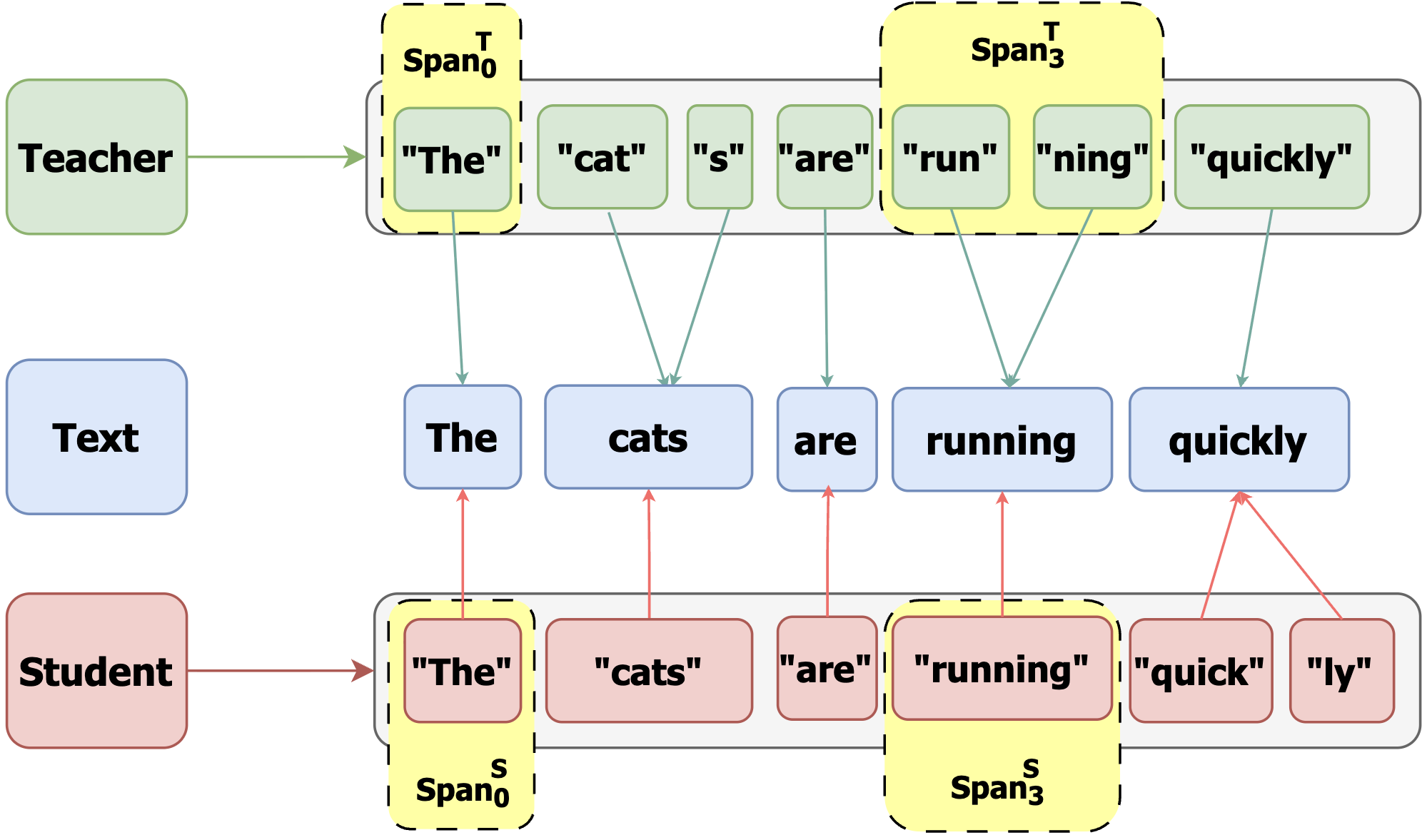}
\captionof{figure}{An illustration of the tokenizer mismatch between a teacher and a student model.}
\label{fig:SRA_overview}
\end{figure}

Grounded in this physical analogy, we introduce \textbf{SRA} (\textbf{S}pan \textbf{R}epresentation \textbf{A}lignment for Large Language Model Distillation), a novel framework that reframes CTKD. Instead of aligning brittle tokens, SRA aligns stable, tokenizer-agnostic spans (Figure~\ref{fig:SRA_overview}). We conceptualize each span as a \textit{cluster of particles} (its constituent tokens), whose state is represented by its \textit{Center of Mass (CoM)} - an attention-weighted average of its particle positions. SRA distills these CoM representations and employs a geometric regularizer to preserve their relative positions in the representation space.

We demonstrate the effectiveness of SRA by transferring knowledge from powerful LLMs \citep{bai2023qwen, jiang2024identifying, yang2024qwen2} into smaller students \citep{radford2019language, zhang2024tinyllama, zhang2022optopenpretrainedtransformer} on multiple benchmarks. SRA consistently outperforms recent CTKD baselines \citep{wan2024knowledgefusionlargelanguage, zhang2024dualspaceknowledgedistillationlarge, cui2025multi, boizard2025crosstokenizerdistillationuniversallogit, chen2025enhancingcrosstokenizerknowledgedistillation}, and our ablations confirm that both the attention-weighted span representations and the geometric regularizer are critical to its success.
In this work, we make the following contributions:
{\sloppy
\begin{enumerate}
\item We introduce a novel conceptual framework for CTKD, grounded in the multi-particle view of Transformers \citep{lu2019understanding}. This allows us to model spans as particle clusters and leverage their \textbf{center of mass (CoM)} for stable, additive representations, moving beyond brittle token-level alignment and naturally supporting the transfer of dynamic, layer-to-layer feature evolution.
\item We propose to distill knowledge at the span level, encompassing both representations and logits. Specifically, we align span representations while preserving their geometric relationships. Furthermore, we introduce an attention-derived weighting scheme inspired by the Center of Mass (CoM) concept, which emphasizes the most salient spans and guides the student to focus on semantically important regions.
\item We evaluate SRA for cross-tokenizer and same tokenizer \textbf{decoder$\rightarrow$decoder} distillation across diverse teacher–student pairs, demonstrating that SRA significantly outperforms recent CTKD baselines on multiple benchmarks. Extensive ablations validate the importance of our core components.
\end{enumerate}
\section{Related Work and Background}
This section reviews prior work in knowledge distillation, focusing on techniques for same- and cross-tokenizer scenario.
\subsection{Related Work}

\paragraph{Knowledge Distillation and Cross-Tokenizer Methods}
Knowledge Distillation (KD) was pioneered by \citep{hinton2015distillingknowledgeneuralnetwork} to transfer knowledge from a large teacher model to a smaller student by matching softened logit distributions. This was later extended to intermediate representations \citep{ sun2019patientknowledgedistillationbert}. These foundational methods and their modern generative counterparts, including highly successful models \citep{gu2024minillmknowledgedistillationlarge, ko2024distillmstreamlineddistillationlarge, le2025token},  all operate under the crucial assumption that the teacher and student share an identical tokenizer.
When this assumption is violated, the tokenizer mismatch presents a significant challenge. One line of research tackles this by seeking an explicit structural alignment between the disparate token sequences, using methods ranging from edit distance \citep{wan2024knowledgefusionlargelanguage, vu2026dwa} to more advanced techniques like entropy-weighted Dynamic Time Warping \citep{chen2025enhancingcrosstokenizerknowledgedistillation} or extending these ideas to preference alignment objectives \citep{truong2026ctpd}. A second, more recent line of work bypasses discrete alignment altogether in favor of distributional or representational alignment. This includes projecting representations into a unified space \citep{zhang2024dualspaceknowledgedistillationlarge} or, more powerfully, using Optimal Transport (OT) to match entire output distributions \citep{boizard2025crosstokenizerdistillationuniversallogit, cui2025multi, hoang2026mcw} and other advanced objectives like approximate likelihood matching \citep{Minixhofer2025ALM}. Related representational alignment ideas have also been explored for embedding model distillation \citep{truong2025emo, an2026mol}. 
While these methods have grown in sophistication, they often operate on the outputs of the tokenization process. They are therefore still fundamentally reliant on aligning sequences of discrete, sometimes brittle, token units or their distributions. 
In contrast, our SRA framework steps back from the token level entirely. It works at a more robust, semantically stable granularity: textual spans. We recover these spans from the raw text using character offsets, avoiding any need for direct token- or vocabulary-level alignment.

A deeper insight into the distillation can be gained by viewing the Transformer as a Multi-Particle Dynamical System (MPDS) \citep{lu2019understanding}. This framework establishes a compelling parallel between the structural components of a Transformer and the dynamics of interacting particles, as outlined in Table~\ref{tab:mpds_transformer_equiv} (see Section ~\ref{sec:Transformers_MPDS}).

\subsection{Background}
\subsubsection{Knowledge Distillation Fundamentals}

Knowledge Distillation (KD) \citep{hinton2015distillingknowledgeneuralnetwork} is a well-established model compression paradigm in which a smaller student network is trained to approximate the behavior of a larger teacher. The learning objective typically augments the standard cross-entropy loss on ground-truth labels with a distillation term:
\begin{equation}
    \mathcal{L} = \mathcal{L}_{CE}(y, z_s) + \lambda \, \mathcal{L}_{KD}(z_t, z_s),
\end{equation}
where $z_t$ and $z_s$ denote the teacher and student logits, and $\lambda$ balances the distillation and supervised losses. The distillation term encourages the student’s distribution to match a softened teacher output. For the distillation loss $\mathcal{L}_{KD}$, standard probabilistic measures such as KL divergence, Jensen–Shannon divergence, and cross-entropy are commonly employed.

\subsubsection{Transformers as a Multi-Particle Dynamic System}
\label{sec:Transformers_MPDS}
A more profound understanding of the knowledge transfer process can be achieved by viewing the Transformer through the lens of a Multi-Particle Dynamical System (MPDS) \citep{lu2019understanding}. This framework draws a powerful analogy between the components of a Transformer and the physics of interacting particles, as summarized in Table~\ref{tab:mpds_transformer_equiv} (Appendix \ref{sec:Transformers_MPDS}). Grounding our design in the multi-particle dynamical system (MPDS) view of Transformers \citep{lu2019understanding}, where token hidden states are particle positions evolving by diffusion (attention) and convection (FFN), we argue: to distill dynamics, one must first preserve state geometry. Building on the MPDS view, \citet{gong2025beyond} pioneered the distillation of feature dynamics, aligning the \textit{trajectory} and first-order derivatives of token representations by pushing them through each model’s LM head and matching in the \emph{vocabulary} (logit) space. While insightful, this work is limited to the same-tokenizer setting. To break the tokenizer barrier, our method treat each span as a cluster of particle, represented by their attention-weighted Center of Mass. We then learn the dynamics of the most salient spans.
\section{Methodology}
\begin{figure*}[t]
    \centering
    \includegraphics[width=0.97\textwidth]{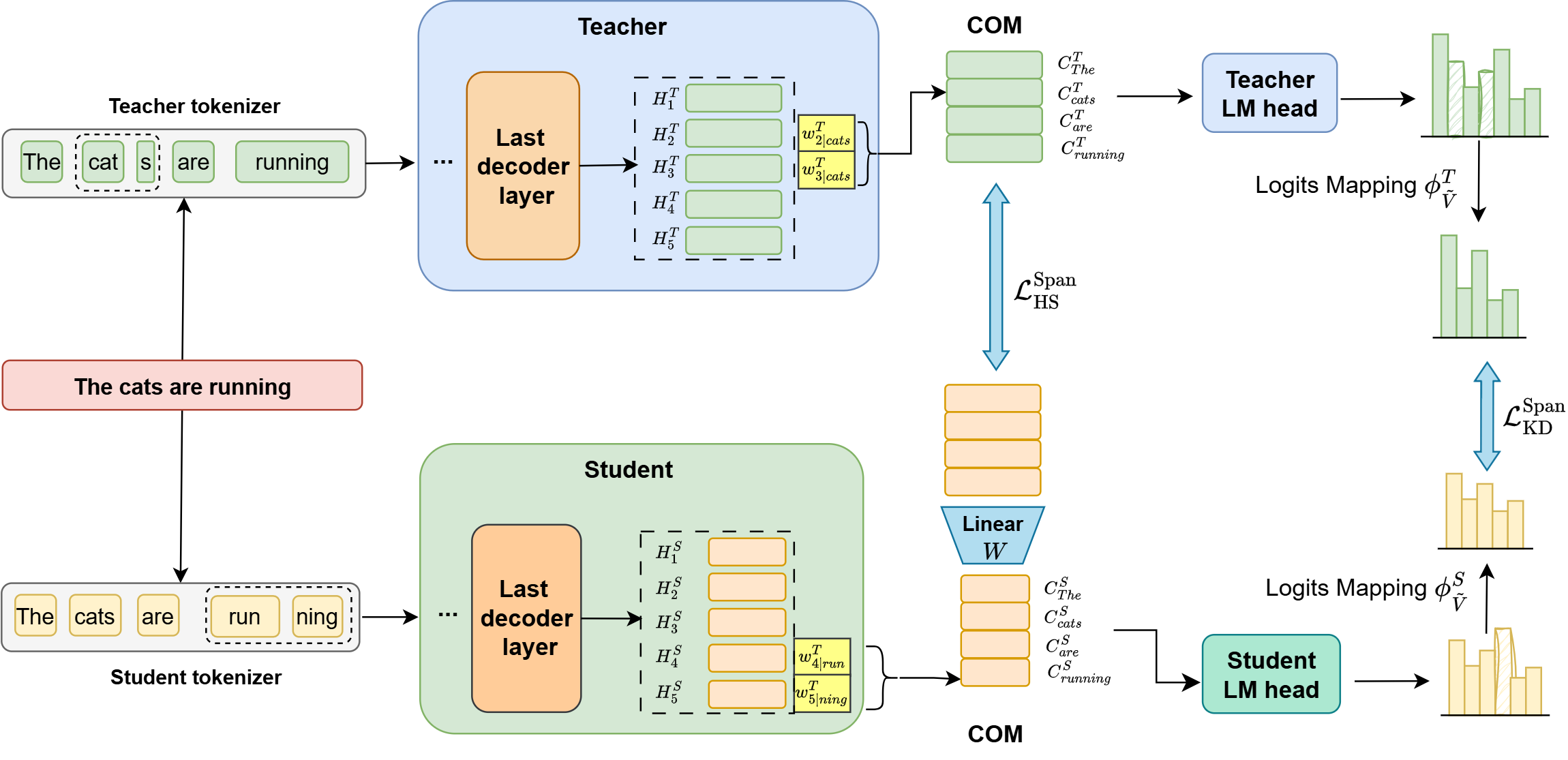} % Full width for A4
    \caption{An illustration of the proposed SRA framework. Teacher–student spans are first matched using longest common subsequence (LCS). Span representations are then obtained via attention-weighted pooling. The student is guided to align its spans with those of the teacher through span-level hidden-state distillation loss ($\mathcal{L}_{HS}^{Span}$) and span-level logits distillation loss ($\mathcal{L}_{KD}^{Span}$).}
    \label{fig:SRE_workflow}
\end{figure*}

In this section, we present \textbf{Span Representation Alignment for Large Language Model Distillation (SRA)}, a framework for knowledge distillation. The core idea of SRA is to construct span-level representations for knowledge transfer, inspired by the center of mass in a multi-particle dynamic system (Section~\ref{sec:SRA}). We transfer span features from the teacher to the student (Sections~\ref{sec:hs_span_loss} and~\ref{sec:span_FD}), allowing the student to better capture the teacher’s structural space in vector representations. The framework SRA is illustrated in Figure~\ref{fig:SRE_workflow}.

\subsection{Span Representation Alignment}
\label{sec:SRA}
Although the teacher and student tokenizers often segment an input sentence into different token sequences, these tokens can still be grouped into spans that correspond to the same textual unit. Direct token-level alignment is therefore infeasible across tokenizers, but span-level alignment provides a natural bridge for transferring knowledge. We use such spans to serve as the basic units of alignment between the teacher and the student. The following subsections describe how spans are aligned across models and how span representations are formed.

\subsubsection*{Span Mapping via Longest Common Subsequence (LCS)}
To construct and align spans, we start from an input sequence $x$ that is tokenized by both the teacher and the student tokenizers. Each tokenizer produces a sequence of tokens together with their character offsets indicating the end position of each token in $x$, i.e., $(t_1, \ldots, t_n)$ with offsets $(\text{off}_1, \ldots, \text{off}_n)$. Applying this process to both models yields two offset sequences $(\text{off}^T_1, \ldots, \text{off}^T_{n_T})$ and $(\text{off}^S_1, \ldots, \text{off}^S_{n_S})$ for the teacher and student, respectively. By computing the LCS of these offset sequences, we identify matched segments of the input that correspond to the same textual unit. These matched segments define the start and end indices of token spans in both models, thereby allowing us to extract pairs of aligned spans from the teacher and student representations. The detailed procedure is described in Appendix~\ref{sec: Appendix_D}.

% \begin{algorithm}[ht]
% \caption{LCS-based Span Alignment}
% \label{alg:span_alignment}
% \small
% \KwIn{Teacher offsets $O^T$, Student offsets $O^S$, \\
% \hspace*{2.85em} First non-special token indices $t^T_0, t^S_0$}

% \KwOut{List of matched span indices $Ms$}\vspace{0.5em}

% $m \gets \text{len}(O^T)$, $n \gets \text{len}(O^S)$\;
% $i \gets 0$, $j \gets 0$, $aligns \gets [(t^T_0, t^S_0)]$, $Ms \gets [\,]$\;

% \While{$i < m$ \textbf{and} $j < n$}{
%     \If{$O^T[i] = 0$}{
%         $i \gets i+1$\;
%         \textbf{continue}\;
%     }
%     \If{$O^S[j] = 0$}{
%         $j \gets j+1$\;
%         \textbf{continue}\;
%     }
%     \eIf{$O^T[i] = O^S[j]$}{
%         $span^S = (aligns[-1][0], i)$\;
%         $span^T = (aligns[-1][1], j)$\;
%         append $(span^S, span^T)$ to $Ms$\;
%         append $(i+1, j+1)$ to $aligns$\;
%         $i \gets i+1$,\, $j \gets j+1$\;
%     }{
%         \eIf{$O^T[i] < O^S[j]$}{
%             $i \gets i+1$\;
%         }{
%             $j \gets j+1$\;
%         }
%     }
% }
% \Return{$Ms$}\;
% \end{algorithm}

We assign special tokens (e.g., [PAD]) an offset value $0$ and then disregard them. Consequently, special tokens are not aligned.
% This guarantees the resulting offset sequence is strictly increasing, ensuring that the subsequence matching can be efficiently computed. 
% As a result, these model-specific tokens, which have different design and usage across models, are correctly excluded from the alignment process.
Compared to token-level alignment, span-level alignment ensures full coverage of the input sentence and mitigates the risk of information loss.

\subsubsection*{Span Representation via Weighted Token Pooling}
While prior work has considered span-based distillation \citep{liu2022multi, chen2025enhancingcrosstokenizerknowledgedistillation}, most approaches adopt simple aggregation strategies such as mean pooling, which risk diluting the contribution of salient tokens. In contrast, we construct span representations through weighted aggregation of token embeddings, supported by a theoretical grounding that ensures more faithful preservation of semantic information. 

% From the perspective of a multi-particle dynamical system, tokens in a sentence can be viewed as particles moving in a $d$-dimensional space. 
% When grouped together, each cluster of particles admits a centroid, which naturally corresponds to the representation of a span. 
% Moreover, the centroid of a larger group can be expressed as a linear combination of the centroids of its subgroups, suggesting that the representation of a span can be constructed as a linear aggregation of its constituent tokens or sub-spans.
The Transformer can be interpreted from a MPDS perspective (Section~\ref{sec:Transformers_MPDS}), tokens in a sentence correspond to particles moving in a $d$-dimensional space, and the hidden states at each layer represent their positions over time. The sequence of hidden states across layers then approximates the trajectories of these particles.

In physics, a center of mass serves as a rigorous representative of a cluster of particles, aggregating the positions and masses of all particles into a single point. A key property of this concept is its hierarchical nature: the center of mass of a larger cluster can be computed as a weighted average of the centers of mass of its subclusters or individual particles (see Appendix~\ref{sec:CoM}). Inspired by the concept of the center of mass, we propose to model the representation of a text span (a cluster of tokens) analogously to the center of mass. This perspective motivates our definition of \textit{a span representation as a linear combination of its constituent token or sub-span representations}. By defining our span representations possess this property, we ensure they can be computed hierarchically from smaller sub-spans.

Based on this principle, instead of aligning individual tokens (particle-level), we align spans (center-of-mass level), thereby transferring features at the span level (Section~\ref{sec:hs_span_loss} and Section~\ref{sec:span_FD}) to address the CTKD problem. This formulation constitutes the core idea of our approach, enabling the model to capture both semantic and structural information while respecting the token-level structure.
Since our knowledge transfer is performed on the last hidden states, we define the token-level weights at the final layer $L$ as:
\begin{equation}
    w_{i} = \frac{\tilde{w}_{i}}{\sum_{t=1}^N \tilde{w}_{t}},
    \quad \tilde{w}_{i} = \sum_{a=1}^{A_h} \text{Att}_{a,N,i}
\label{eq:token_weight}
\end{equation}
where $A_h$ is the number of attention heads, $N$ is the sequence length, and $\text{Att}_{a,N,i}$ denotes the attention score from last token $N$ to token $i$ in head $a$ at layer $L$. As the last token in an LLM typically encodes the global contextual information of the entire sequence, $\tilde{w}_{i}$ quantifies the importance of token $i$ via the attention it receives from the last token aggregated across all heads. 

Given the aligned spans identified by $M\!s = [(Span^S_0, Span^T_0), \dots,(Span^S_n, Span^T_n)]$, with $Span^S_i = (s^S_i, e^S_i)$ and $Span^T_i = (s^T_i, e^T_i)$, where $s$ and $e$ denote the start and end token indices of a span, the span representations at the final layer for the student and teacher models are then computed as:
\begin{equation}
    C^S_i = \sum_{t=s^S_i}^{e^S_i} w^S_{t} \, H^S_{t}
    \label{eq:span_s}
    \end{equation} 
    \begin{equation}
            C^T_i = \sum_{t=s^T_i}^{e^T_i} w^T_{t} \, H^T_{t}
\label{eq:span_t}
\end{equation}
where $H^S_{t}$ and $H^T_{t}$ denote the last hidden representations of token $t$ in the student and teacher, respectively.

\subsection{Span-Level Hidden State Transfer}
\label{sec:hs_span_loss}

Under the lens of Transformers as a Multi-Particle Dynamical System, knowledge distillation can be viewed as transferring the salient characteristics of particles, where each particle’s position corresponds to a token’s hidden state. Consequently, distilling feature positions amounts to aligning the hidden states between teacher and student, consistent with prior work demonstrating the effectiveness of hidden-state distillation~\citep{sanh2020distilbertdistilledversionbert,DBLP:journals/corr/abs-1909-10351}.

However, most existing methods treat all tokens as equal contributors to the distillation loss. From the perspective of the \emph{center of mass} (CoM), this assumption is overly restrictive. In a particle system, each particle has a mass, and heavier particles pull the center of mass closer to themselves, reflecting their greater contribution to the global dynamics. Analogously, tokens vary in semantic importance: highly attended tokens act as ``heavier'' particles that should dominate the representation of the sentence, while less informative tokens contribute less.  

Extending this analogy from tokens to spans, we argue that uniform weighting of spans is suboptimal, as it forces the student to overfit to irrelevant signals and dilutes the transfer of meaningful knowledge from teacher to student. Instead, we propose to assign attention-derived weights to spans, reflecting their semantic salience in the teacher model. Building on this intuition, we formulate a weighted hidden-state transfer loss, $\mathcal{L}_\text{HS}^\text{Span}$, that emphasizes the most informative spans while down-weighting unimportant ones: 

\begin{equation}
    \mathcal{L}_\text{HS}^\text{Span} = \sum_{i=1}^{N_{sp}} w^\text{sp}_{i}\ \mathcal{L}_{\text{cos}} \! \left(C^S_i W, C^T_i\right) + \lambda \mathcal{L}_\text{Geo}
\end{equation}

where $N_{sp}$ is the number of aligned spans, $\mathcal{L}_{\cos}(u,v) = 1 - \frac{u \cdot v}{\|u\| \, \|v\|}$, $C^S_i$ and $C^T_i$ are the student and teacher representations of span $i$ from the final layer (Eq.~\ref{eq:span_s}, Eq.~\ref{eq:span_t}), and $W$ is a learnable projection matrix. The coefficient $\lambda$ is a balancing hyperparameter that controls the strength of the geometric regularizer ($\mathcal{L}_\text{Geo}$).

Unconstrained linear projections can distort the geometry of representations and misalign their relative positions~\citep{miles2024vkd}. To mitigate this issue, we introduce a geometric structure regularizer $\mathcal{L}_\text{Geo}$ that preserves relative geometry between spans:

\begin{equation}
% \small
    \mathcal{L}_\text{Geo} = \sum_{i=1}^{N_{sp}} \! \sum_{j=i+1}^{N_{sp}} w^\text{sc}_{i, j} \! \left(d(C_i^S \!,C_j^S) - d(C_i^T \!,C_j^T)\right)^2
\end{equation}
with
\begin{equation}
    w^\text{sc}_{i, j} = \frac{w^\text{sp}_{i} w^\text{sp}_{j}}{\sum_{i=1}^{N_{sp}} \sum_{j=i+1}^{N_{sp}}w^\text{sp}_{i} w^\text{sp}_{j}}
\end{equation}
where $d(\cdot, \cdot)$ is the cosine function. The normalization ensures $\sum_{i < j} w^\text{sc}_{i, j} = 1$, making the loss comparable across samples. Preserving relative positions is essential for faithful knowledge transfer under the multi-particle perspective.

Reusing the teacher’s token weight $w^T_t$ (Eq.~\ref{eq:token_weight}), we define the normalized span weight as:
\begin{equation}
    w^\text{sp}_{i} = \frac{\tilde{w}^\text{sp}_{i}}{\sum_{t=1}^{N_{sp}} \tilde{w}^\text{sp}_{t}}, \quad \tilde{w}^\text{sp}_{i} = \left( \sum_{t=s^T_i}^{e^T_i} \, w^T_t \right)^p
\label{eq:sp_w}
\end{equation}
where $s^T_i$ and $e^T_i$ are the start and end token indices of the $i$-th teacher span. The hyperparameter $p$ controls the sharpness of the weight distribution; when $p=0$, all spans are weighted equally.

% This loss effectively guides the student to track the teacher’s span-level feature trajectory, while respecting the inherent importance of different spans. 

\subsection{Span-Level Logits Transfer}
\label{sec:span_FD}
 
We incorporate logits distillation into span-level knowledge transfer. Unlike previous approaches such as DSKD~\citep{zhang2024dualspaceknowledgedistillationlarge} or MINED~\citep{wan2024knowledgefusionlargelanguage} , which require complex 
token-level logits alignment, we take a multi-particle perspective. 
From this view, the logits of each token can be regarded as the position 
of a particle projected into the vocabulary space. Consequently, token-level 
logits can be naturally aggregated into span-level logits, enabling more 
faithful and robust knowledge transfer. 

The primary challenge of vocabulary mismatch ($V_{\text{tea}} \neq V_{\text{stu}}$) is then interpreted as the teacher and student particles residing in distinct spaces. Our solution is to perform knowledge transfer within the shared subspace $\tilde{V} = V_{\text{tea}} \cap V_{\text{stu}}$.  This shared subspace represents the overlapping dimensions where particle positions are directly comparable, allowing lexical knowledge to be meaningfully transferred.
We formally define the aligned span logits as:
\begin{equation}
    \tilde{y}_{1:N_{sp}}^S = \phi_{\tilde{V}}^S \left(f_{head}^S(C_{1:N_{sp}}^S)\right)
\end{equation}

\begin{equation}
    \tilde{y}_{1:N_{sp}}^T = \phi_{\tilde{V}}^T \left(f_{head}^T(C_{1:N_{sp}}^T)\right)
\end{equation}

where $\phi_{\tilde{V}}^S(\cdot)$ and $\phi_{\tilde{V}}^T(\cdot)$ are mapping functions projecting the student and teacher logits into the shared subspace $\tilde{V}$, and $f_{head}^S$, $f_{head}^T$ are the LM head layers.

Building on these definitions, we implement the aligned span logits loss, which aligns the distributions of the teacher’s and student’s aligned span logits. We then define the loss as:
\begin{equation}
    \mathcal{L}_\text{KD}^\text{Span} = \sum_{i = 1}^{N_{sp}} \mathcal{L}_{\text{KL}} \left(\tilde{y}_i^T, \tilde{y}_i^S; \tau \right) 
\end{equation}
where $\mathcal{L}_{\text{KL}}(\tilde{y}_i^T, \tilde{y}_i^S; \tau)$ denotes the Kullback–Leibler divergence between the teacher’s and student’s aligned span logits after applying a temperature-scaled ($\tau$) softmax.

\subsection{Combined Distillation Objective}

The overall training objective in SRA integrates complementary span-level 
distillation signals. First, the span-level hidden state loss 
$\mathcal{L}_\text{HS}^\text{Span}$ encourages the student to imitate the teacher's span representations, which are weighted by semantic importance. This loss also incorporates a geometric structure regularizer, $\mathcal{L}_\text{Geo}$, to preserve the relative relationships between spans and prevent distortion of the learned space. 
Second, the aligned span logits loss $\mathcal{L}_\text{KD}^\text{Span}$ ensures knowledge transfer in the vocabulary space by aligning predictive distributions restricted to the shared subspace $\tilde{V}$. 
To further ensure effective learning, we balance these distillation signals with the task-specific supervision from ground-truth labels.
Formally, the overall loss $\mathcal{L}_{\text{overall}}$ is defined as:

\begin{equation}
    \mathcal{L}_{\text{overall}} = \alpha \, \mathcal{L}_{\text{CE}} + (1-\alpha) \, (\mathcal{L}_\text{HS}^\text{Span} + \mathcal{L}_\text{KD}^\text{Span}) 
\end{equation}

where $\alpha \in [0,1]$ balances the contribution of the standard cross-entropy loss $\mathcal{L}_{\text{CE}}$ and the span representation distillation loss.

\section{Experiments}

\subsection{Experimental Setup}
\paragraph{Datasets.} Our evaluation spans multiple instruction-following datasets. We adopt the preprocessing procedure of \citep{wan2024knowledgefusionlargelanguage}. Distillation is trained on \textsc{Databricks-Dolly-15k}. For evaluation, we report ROUGE-L on the Dolly test split and on four benchmarks - \textsc{S-NI} \citep{wang2022benchmarking}, \textsc{VicunaEval} \citep{chiang2023vicuna}, \textsc{DialogSum} \citep{chen2021dialogsumreallifescenariodialogue}, and \textsc{SelfInst} \citep{wang2023selfinstructaligninglanguagemodels} - providing a broad view of models’ cross-domain generalization.
\paragraph{Training and Evaluation Settings.} Our experiments target cross-tokenizer distillation where teacher and student use different vocabularies. Students span GPT-2 (120M, 340M, 1.5B) \citep{radford2019language}, TinyLLaMA-1.1B \citep{zhang2024tinyllama}, and OPT-2.7B \citep{zhang2022optopenpretrainedtransformer}. We evaluate the following teacher$\rightarrow$student pairs: Qwen1.5–1.8B \citep{bai2023qwen} $\rightarrow$ GPT-2-120M / GPT-2-340M; Qwen2.5-7B-Instruct \citep{yang2024qwen2} $\rightarrow$ GPT-2-1.5B / OPT-2.7B; Mistral-7B \citep{jiang2024identifying}$\rightarrow$ TinyLLaMA-1.1B; and GPT-2-1.5B $\rightarrow$ GPT-2-120M. Training and evaluation setup, and details of baseline models are provided in Appendix~\ref{sec: appendix_B}. Reported evaluation results are averaged over five random seeds.

\paragraph{Baselines.}
We compare \textbf{SRA} with current CTKD methods \footnote[1]{
We exclude CDM~\citep{chen2025enhancingcrosstokenizerknowledgedistillation} from our baselines. Although we attempted to implement it, the per-step computational cost—driven by its dynamic programming alignment for each training batch (as shown in Table \ref{tab:efficiency})—proved to be substantially more expensive and 
impractical for our large-scale experimental setup.
}: ULD \citep{boizard2025crosstokenizerdistillationuniversallogit}, DSKD \citep{zhang2024dualspaceknowledgedistillationlarge}, MinED \citep{wan2024knowledgefusionlargelanguage} and MultilevelOT \citep{cui2025multi}. 

\subsection{Main Results}

\begin{table}[h!]
\centering
\footnotesize
\renewcommand{\arraystretch}{1.15}
\setlength{\tabcolsep}{1.5pt}

\begin{tabular}{l|cccccc}
\toprule
\textbf{Methods} & \textbf{Dolly} & \textbf{Vicuna} & \textbf{SelfInst} & \textbf{S-NI} & \textbf{Dialog} & \textbf{Avg.}\\
\specialrule{1.0pt}{1.0pt}{1.0pt}

% ================= Qwen1.5 -> GPT-2 120M =================
\multicolumn{7}{c}{\textit{Qwen1.5--1.8B} $\rightarrow$ \textit{GPT-2 120M}}\\
\midrule
Teacher        & 28.23 & 19.59 & 19.58 & 34.36 & 14.18 & 23.19 \\
\midrule
SFT            & 23.78 & 17.04 &  6.78 &  7.81 &  8.29 & 12.74 \\
ULD            & 23.77 & 14.33 &  9.30 & 14.04 &  8.63 & 14.01 \\
MinED          & 24.21 & 14.96 & 10.02 & 16.40 &  9.79 & 15.08 \\
MultiLevelOT   & 23.02 & 13.79 &  8.41 & 12.26 &  8.79 & 13.25 \\
DSKD           & \textbf{24.26} & 15.25 & 10.07 & 17.15 & 10.03 & 15.35 \\
\rowcolor{gray!20}
\textbf{SRA}   & 23.36 & \textbf{16.08} & \textbf{13.16} & \textbf{23.70} & \textbf{13.56} & \textbf{17.97} \\
\specialrule{1.0pt}{1.0pt}{1.0pt}

% ================= Qwen1.5 -> GPT-2 340M =================
\multicolumn{7}{c}{\textit{Qwen1.5--1.8B} $\rightarrow$ \textit{GPT-2 340M}}\\
\midrule
Teacher        & 28.23 & 19.59 & 19.58 & 34.36 & 14.18 & 23.19 \\
\midrule
SFT            & 23.11 & 14.89 &  9.09 & 13.03 &  8.00 & 13.62 \\
ULD            & 23.90 & 15.04 &  9.96 & 16.26 &  8.76 & 14.78 \\
MinED          & 24.48 & 15.56 & 11.21 & 15.69 &  8.98 & 15.18 \\
MultiLevelOT   & 23.95 & 14.80 & 10.21 & 15.87 &  8.99 & 14.76 \\
DSKD           & \textbf{25.43} & 15.08 & 11.29 & 17.18 &  8.90 & 15.57 \\
\rowcolor{gray!20}
\textbf{SRA}   & 23.97 & \textbf{16.31} & \textbf{13.64} & \textbf{24.49} & \textbf{12.08} & \textbf{18.10} \\
\specialrule{1.0pt}{1.0pt}{1.0pt}

% ================= Qwen2.5 -> GPT2-1.5B =================
\multicolumn{7}{c}{\textit{Qwen2.5-7B-Instruct} $\rightarrow$ \textit{GPT2-1.5B}}\\
\midrule
Teacher        & 28.49 & 20.48 & 24.67 & 39.87 & 16.86 & 26.07 \\
\midrule
SFT            & 21.83 & 15.95 & 13.62 & 21.66 & 10.91 & 16.79 \\
ULD            & 24.52 & 15.94 & 15.11 & 26.18 & 11.72 & 18.69 \\
MinED          & 25.52 & 16.15 & 15.39 & 26.25 & 11.79 & 19.02 \\
MultilevelOT   & 24.40 & 15.97 & 14.53 & 23.94 & 10.84 & 17.94 \\
DSKD           & 25.38 & 16.84 & 16.10 & 25.82 & 12.19 & 19.27 \\
\rowcolor{gray!20}
\textbf{SRA}   & \textbf{26.45} & \textbf{18.25} & \textbf{17.22} & \textbf{29.50} & \textbf{13.52} & \textbf{20.99} \\
\specialrule{1.0pt}{1.0pt}{1.0pt}

% ================= Qwen2.5 -> OPT-2.7B =================
\multicolumn{7}{c}{\textit{Qwen2.5-7B-Instruct} $\rightarrow$ \textit{OPT-2.7B}}\\
\midrule
Teacher        & 28.49 & 20.48 & 24.67 & 39.87 & 16.86 & 26.07 \\
\midrule
SFT            & 27.10 & 16.60 & 13.90 & 24.90 & 10.62 & 18.62 \\
ULD            & 26.65 & 16.97 & 15.37 & 25.44 & 12.15 & 19.32 \\
MinED          & 26.89 & 17.04 & 14.98 & 25.94 & 11.78 & 19.33 \\
MultilevelOT   & 26.76 & 16.56 & 15.51 & 24.84 & 11.43 & 19.02 \\
DSKD           & 26.93 & 17.86 & 16.22 & 27.33 & 12.43 & 20.15 \\
\rowcolor{gray!20}
\textbf{SRA}   & \textbf{28.52} & \textbf{18.48} & \textbf{17.14} & \textbf{27.35} & \textbf{13.13} & \textbf{20.92} \\
\specialrule{1.0pt}{1.0pt}{1.0pt}

% ================= Mistral -> TinyLLaMA =================
\multicolumn{7}{c}{\textit{Mistral-7B} $\rightarrow$ \textit{TinyLLaMA-1.1B}}\\
\midrule
Teacher        & 32.15 & 20.43 & 25.44 & 36.88 & 14.67 & 25.91 \\
\midrule
SFT            & 23.20 & 15.70 & 15.70 & 28.43 & 10.77 & 18.76 \\
ULD            & 25.48 & 17.31 & 17.72 & 32.54 & 11.75 & 20.96 \\
MinED          & 25.54 & 17.02 & 18.23 & 31.42 & 11.77 & 20.80 \\
MultilevelOT   & 24.56 & 16.84 & 15.61 & 27.91 & 12.04 & 19.40 \\
DSKD           & \textbf{26.28} & 18.74 & 17.19 & 31.93 & 12.53 & 21.33 \\
\rowcolor{gray!20}
\textbf{SRA}   & 25.02 & \textbf{19.69} & \textbf{20.05} & \textbf{32.98} & \textbf{14.88} & \textbf{22.52} \\
\bottomrule
\end{tabular}
\caption{Performance comparison of different teacher–student model combinations for CTKD.}
\label{tab:main_results}
\end{table}

Table~\ref{tab:main_results} provides an overview of the ROUGE-L evaluation results for all teacher–student configurations examined in this study. The first two sections correspond to medium-size teacher–student pairs, while the last three sections summarize the results for larger models. Across all teacher–student pairs and evaluation benchmarks, SRA consistently outperforms strong CTKD baselines (ULD, MinED, DSKD, and MultilevelOT), achieving the highest scores on most datasets and the best average ROUGE-L in every setting. These results highlight the effectiveness of span-level alignment in narrowing the performance gap between student and teacher models. We additionally compare SRA against ALM~\citep{Minixhofer2025ALM},
a span-based cross-tokenizer distillation method; as shown in 
Appendix~\ref{app:alm}, SRA consistently outperforms ALM 
across all benchmarks and representative teacher–student configurations.

\begin{figure}[h!]
    \centering
    \includegraphics[width=1\linewidth]{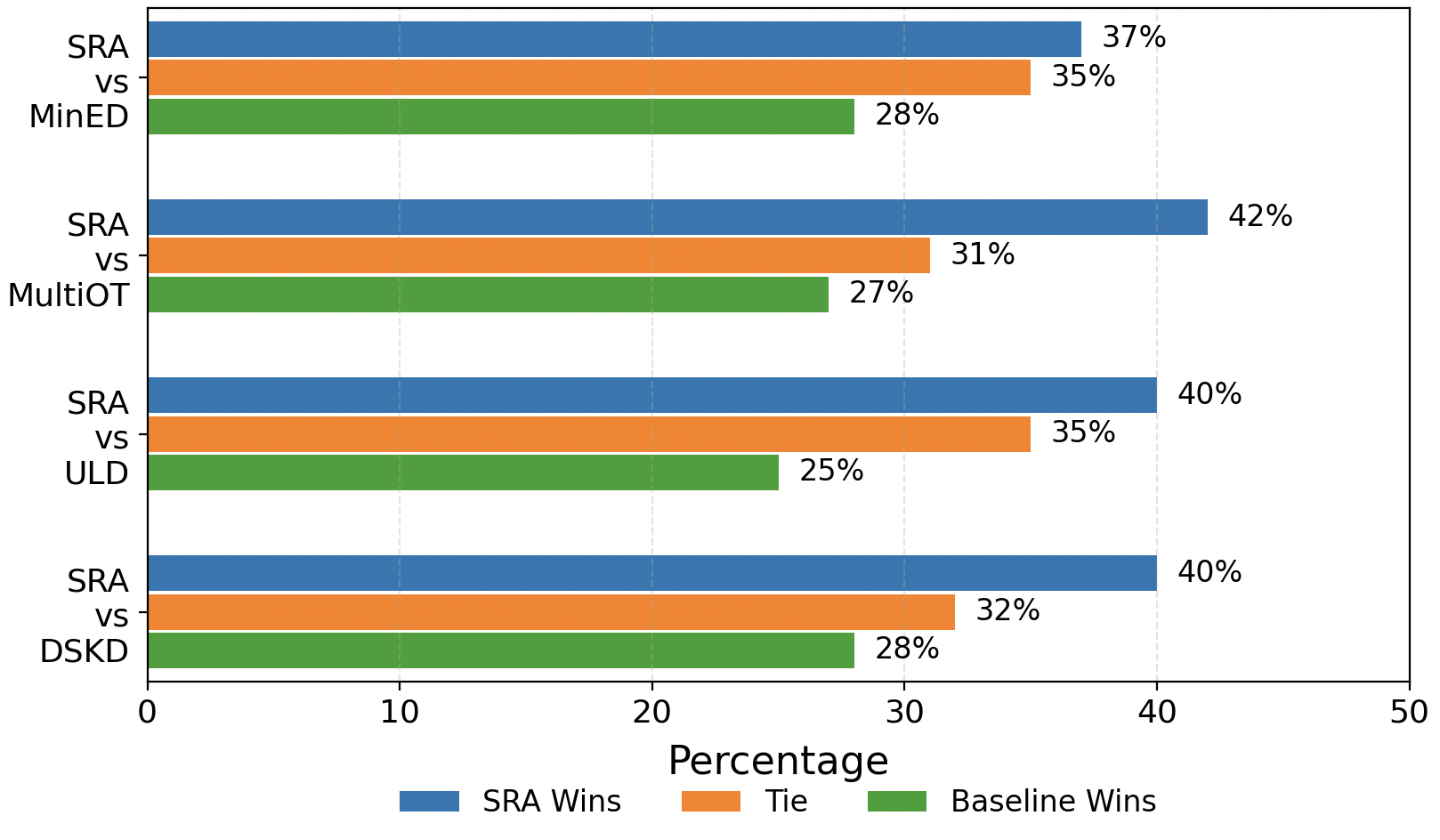}
    \caption{Win rates (\%) for distilling Qwen 2.5-7B$\rightarrow $GPT2 1.5B, evaluated by GPT-4o-mini.}
    \label{fig:placeholder1}
\end{figure}
% \begin{figure}[h!]
%     \centering
%     \includegraphics[width=1\linewidth]{gpt4.1 eval1.png}
%     \caption{Caption}
%     \label{fig:placeholder2}
% \end{figure}
% \subsection{Ablation Study}

Figure~\ref{fig:placeholder1} reports the results of the semantic evaluation. To compare SRA with each baseline, we randomly sampled data from the combined benchmark corpus and employed \texttt{GPT-4o-mini} as an automatic judge. The model was prompted to determine which system’s output was semantically superior by considering multiple criteria, including helpfulness, relevance and accuracy when responding to diverse instructions. Complementing the quantitative results in the previous table, these findings further confirm that SRA consistently delivers more semantically faithful and contextually aligned responses, demonstrating its strong robustness and generalization capability across different tasks.

\section{Analysis}
This section further demonstrates the contribution of each component in our method. Additional ablation studies are provided in Appendix~\ref{sec: Appendix_C}.
\paragraph{Impact of Span-Level Hidden State Transfer}
Table~\ref{tab:loss_ablation_if} shows that transferring the last-layer hidden state knowledge enhances the generative ability of the student models. Both the geometric loss ($\mathcal{L}_{\text{Geo}}$) and the cosine loss ($\mathcal{L}_{\text{cos}}$) contribute individual gains. The geometric loss gives the largest gain, and the cosine loss, though minor, further improves results when combined. This confirms the effectiveness of SRA’s design, where complementary loss terms jointly strengthen cross-tokenizer knowledge transfer.

\paragraph{Ablation on Weighting Mechanisms}
Table~\ref{tab:span_hs_ablation} further isolates the effect of Weighted Span Pooling (WSP) and Weighted Span Loss (WSL). Removing WSP is equivalent to applying mean pooling over spans, while ablating WSL treats all tokens as equally important in the span-level objective. These components play a crucial role in forming high-quality span representations prior to execution of the method. The results show that removing either term weakens span representations and leads to less effective knowledge transfer, ultimately degrading distillation performance. This highlights the importance of well-structured span-level hidden state transfer in the overall effectiveness of SRA.

\paragraph{Results with shared vocabulary}
Table ~\ref{tab:same_tokenizer} demonstrates that SRA outperforms two categories of existing methods. The first includes distance-based techniques for shared-tokenizer scenarios, such as SeqKD~\citep{kim2016sequencelevelknowledgedistillation}, reverse KL (RKL) and Jensen–Shannon (JS) divergence~\citep{wen2023fdivergenceminimizationsequencelevelknowledge}, skewed KL (SKL) and skewed reverse KL (SRKL) divergence~\citep{ko2024distillmstreamlineddistillationlarge}, and adaptive KL divergence (AKL)~\citep{wu2024rethinkingkullbackleiblerdivergenceknowledge}. The second includes methods addressing tokenizer discrepancies, such as ULD~\citep{boizard2025crosstokenizerdistillationuniversallogit}, MinED~\citep{wan2024knowledgefusionlargelanguage}, DSKD~\citep{zhang2024dualspaceknowledgedistillationlarge}, and MultilevelOT~\citep{cui2025multi}.
SRA yields higher ROUGE-L scores across multiple datasets and the highest overall average, despite being primarily designed to handle tokenizer discrepancies. This improvement can stem from SRA’s ability to capture diverse representational differences between teacher and student, leading to more stable and comprehensive knowledge transfer.

\begin{table}[h!]
\centering
\scriptsize
\renewcommand{\arraystretch}{1.00}
\setlength{\tabcolsep}{1.7pt}
\begin{tabular}{ccc|cccccc}
\toprule
$\mathcal{L}_{KD}^{Span}$ & $\mathcal{L}_{Geo}$ & $\mathcal{L}_{cos}$ & \textbf{Dolly} & \textbf{Vicuna} & \textbf{SelfInst} & \textbf{S-NI} & \textbf{Dialog} & \textbf{Avg} \\
\specialrule{1.0pt}{1.0pt}{1.0pt}
\multicolumn{9}{c}{\textit{Qwen1.5--1.8B} $\rightarrow$ \textit{GPT-2 340M}} \\
\midrule
\checkmark &            &            & 22.98 & \textbf{16.32} & 11.76 & 24.10 & \underline{11.64} & 17.36 \\
\checkmark & \checkmark &            & 23.58 & 16.05 & \textbf{13.73} & \textbf{25.02} & 11.32 & \underline{17.94} \\
\checkmark &            & \checkmark & 23.09 & 16.23 & 12.76 & \underline{24.78} & 10.84 & 17.54 \\
           & \checkmark & \checkmark & \textbf{25.56} & 15.93 & 13.28 & 23.70 &  8.95 & 17.48 \\
\checkmark & \checkmark & \checkmark & \underline{23.97} & \underline{16.31} & \underline{13.64} & 24.49 & \textbf{12.08} & \textbf{18.10} \\
\specialrule{1.0pt}{1.0pt}{1.0pt}
\multicolumn{9}{c}{\textit{Qwen1.5--1.8B} $\rightarrow$ \textit{GPT-2 120M}} \\
\midrule
\checkmark &            &            & \textbf{23.74} & 15.85 & 12.03 & 22.11 & 11.78 & 17.10 \\
\checkmark & \checkmark &            & 22.80 & 15.96 & \underline{12.94} & \textbf{25.05} & 11.85 & \underline{17.72} \\
\checkmark &            & \checkmark & 23.20 & \textbf{16.16} & 12.02 & 22.55 & \underline{12.69} & 17.32 \\
           & \checkmark & \checkmark & 23.24 & 15.18 & 11.02 & 21.45 &  9.32 & 16.04 \\
\checkmark & \checkmark & \checkmark & \underline{23.36} & \underline{16.08} & \textbf{13.16} & \underline{23.70} & \textbf{13.56} & \textbf{17.97} \\
\bottomrule
\end{tabular}
\caption{Ablation study on loss components across two teacher--student pairs. 
\textbf{Bold} denotes the highest score and \underline{underline} the second highest.}
\label{tab:loss_ablation_if}
\end{table}

% Requires: \usepackage{booktabs}
\begin{table}[h!]
\centering
\footnotesize% hoặc \scriptsize nếu vẫn còn rộng
\setlength{\tabcolsep}{1pt}     % thu hẹp khoảng cách giữa các cột (mặc định ~6pt)
\renewcommand{\arraystretch}{1.0} % điều chỉnh độ cao hàng (1.0 là mặc định)
\begin{tabular}{c c | c c c c c | c}
\toprule
WSL & WSP &
\textbf{Dolly} & \textbf{Vicuna} & \textbf{SelfInst} & \textbf{S-NI} & \textbf{Dialog} & \textbf{Avg} \\
\midrule
\multicolumn{8}{c}{\textit{Qwen1.5-1.8B $\rightarrow$ GPT-2 340M}} \\
\midrule
            &            & 22.49 & 14.25 & \underline{12.90} & \underline{23.30} & 12.02 & 16.99    \\ % w/o both
            & \checkmark & 23.18 & \underline{15.57} & 12.82 & 21.99 & 12.01 & 17.11    \\ % w/o span loss weight
\checkmark  &            & \underline{23.25} & 15.43 & 12.82 & 22.35 & \textbf{12.93} &                 \underline{17.36}   \\ % w/o span loss pooling
\checkmark  & \checkmark & \textbf{23.97} & \textbf{16.31} & \textbf{13.64} & \textbf{24.49} & \underline{12.08} & \textbf{18.10} \\ % normal
\midrule
\multicolumn{8}{c}{\textit{Qwen1.5-1.8B $\rightarrow$ GPT-2 120M}} \\
\midrule
            &            & 20.12 & 12.23 & 11.50 & 19.31 & 11.07 & 14.85    \\ % w/o both
            & \checkmark & 20.86 & 13.16 & \underline{12.36} & \underline{20.97} & 11.52 & 15.77    \\ % w/o span loss weight
\checkmark  &            & \underline{21.54} & \underline{13.49} & 11.09 & 20.69 & \underline{12.62} &                 \underline{15.89}    \\ % w/o span loss pooling
\checkmark  & \checkmark & \textbf{23.36} & \textbf{16.08} & \textbf{13.16} & \textbf{23.70} & \textbf{13.56} & \textbf{17.97} \\ % normal
\bottomrule
\end{tabular}
\caption{Ablations on weighting mechanisms. Weighted Span Loss (WSL); Weighted Span Pooling (WSP).}
\label{tab:span_hs_ablation}
\end{table}

\begin{table}[h!]
\centering
% \scriptsize
\footnotesize
\renewcommand{\arraystretch}{1.0}
\setlength{\tabcolsep}{2pt} 

\begin{tabular}{l|ccccc}
\toprule
\textbf{Methods} & \textbf{Dolly} & \textbf{Vicuna} & \textbf{SelfInst} & \textbf{S-NI} & \textbf{Avg.} \\
\specialrule{1.0pt}{1.0pt}{1.0pt}
\multicolumn{6}{c}{\textit{GPT-2-1.5B} $\rightarrow$ \textit{GPT-2--120M}}\\
\midrule
Teacher       & 27.19 & 16.30 & 14.64 & 27.55 & 21.42 \\
\midrule
SFT           & 22.94 & 15.17 & 10.11 & 16.21 & 16.11 \\
SeqKD         & 23.68 & 14.41 & 10.03 & 16.61 & 16.18 \\
RKL           & 24.34 & 15.71 & 10.53 & 17.31 & 17.03 \\
JS            & 23.86 & 15.50 & 10.20 & 16.20 & 16.44 \\
SKL           & 24.03 & 14.70 & 10.66 & 17.99 & 16.85 \\
SRKL          & 23.48 & 14.91 & 10.35 & 16.53 & 16.32 \\
AKL           & \textbf{24.75} & 15.37 & 10.46 & 17.48 & 17.02 \\
ULD           & 23.53 & 14.89 & 10.47 & 15.43 & 16.08 \\
MinED         & 23.69 & 15.17 & 10.43 & 15.84 & 16.28 \\
MultiLevelOT  & 23.81 & 14.91 & 10.70 & 14.91 & 16.08 \\
DSKD          & 23.93 & 15.00 & 10.66 & 16.81 & 16.60 \\
\rowcolor{gray!20}
\textbf{SRA} & 23.21 & \textbf{16.17} & \textbf{12.53} & \textbf{25.06} & \textbf{19.24}  \\
\bottomrule
\end{tabular}
\caption{Experiment on same tokenizer distillation.}
\label{tab:same_tokenizer}
\end{table}

\paragraph{Training Efficiency}

\begin{table}[h!]
\centering
\footnotesize
\renewcommand{\arraystretch}{1.15}
\setlength{\tabcolsep}{6pt}
\begin{tabular}{l|cc}
\toprule
\textbf{Method} & \textbf{avg\_alloc (GB)} & \textbf{avg\_step\_time (s)} \\
\midrule
CDM$^\dagger$ & 22.61 & 1.0100 \\
DSKD  & 20.11 & 0.3520 \\
MinED & 19.63 & 0.4244 \\
ULD   & 19.63 & 0.4393 \\
\rowcolor{gray!20}
\textbf{SRA} & 21.96 & \textbf{0.2754} \\
\bottomrule
\end{tabular}
\caption{Average GPU memory allocation and per-step training time for 
Qwen2.5-7B-Instruct $\rightarrow$ GPT2-1.5B on a single A100 40GB GPU 
(200 steps). DSKD, MinED, ULD, and SRA use batch size = 4; 
$^\dagger$CDM is run with batch size = 1 due to memory constraints.}
\label{tab:efficiency}
\end{table}

Table~\ref{tab:efficiency} reports the training efficiency of \textbf{SRA} 
compared to CTKD baselines under identical settings. Although SRA requires 
slightly more GPU memory than MinED and ULD ($\approx$1--1.5 GB on a 40 GB 
card), owing to the additional span matching and attention-based aggregation 
components, it achieves the \textbf{lowest average per-step training time} 
among all methods. In other words, the span-level geometric terms introduce only a modest memory increase while keeping the overall training cost comparable to, and in this setting even more time-efficient than, strong CTKD baselines. 

\paragraph{Robustness to Vocabulary Mismatch} Table~\ref{tab:vocab_overlap} reports the vocabulary overlap between 
each teacher–student pair used in our experiments. In all configurations, 
the shared vocabulary covers 76--84\% of the student's vocabulary, 
ensuring that span-level logit distillation operates over a substantial 
and semantically meaningful subset of the lexical space.
Notably, even without logit distillation loss ($L_{KD}^{Span}$), the span-level hidden-state alignment alone ($\mathcal{L}_{cos}$ + 
$\mathcal{L}_{Geo}$) already yields competitive performance against strong CTKD 
baselines (see in Appendix \ref{sec:sra_wo_logit}), demonstrating that our core span representations are 
inherently tokenizer-agnostic. However, incorporating the logit distillation 
loss $\mathcal{L}_{cos}$ over the shared vocabulary further boosts performance 
across all configurations (see Table~\ref{tab:loss_ablation_if}).
\begin{table}[h!]
\centering
\scriptsize
\renewcommand{\arraystretch}{1.15}
\setlength{\tabcolsep}{3pt}
\begin{tabular}{l|l|cccc}
\toprule
Teacher & Student & \textbf{|V$_S$|} & \textbf{|V$_T$|} & \textbf{|V$_S \cap$ V$_T$|} & \textbf{\%} \\
\midrule
Qwen1.5-1.8B & GPT-2 120M   & 50,257 & 151,646 & 42,257 & 84 \\
Qwen1.5-1.8B & GPT-2 340M   & 50,257 & 151,646 & 42,257 & 84 \\
Qwen2.5-7B & GPT2-1.5B      & 50,257 & 151,665 & 42,257 & 84 \\
Qwen2.5-7B & OPT-2.7B       & 50,265 & 151,665 & 42,260 & 84 \\
Mistral-7B & TinyLLaMA-1.1B & 32,000 &  32,000 & 24,184 & 76 \\
\bottomrule
\end{tabular}
\caption{Vocabulary overlap between teacher and student tokenizers. 
\% denotes the fraction of the student vocabulary present in the teacher vocabulary.}
\label{tab:vocab_overlap}
\end{table}

\section{Conclusion}
We introduced \textbf{SRA}, a framework for CTKD that leverages span-level alignment to bridge the gap between heterogeneous tokenizations. By aligning sequences via LCS, SRA provides a principled way to construct comparable span representations while capturing span importance. Extensive experiments across multiple benchmarks demonstrate that SRA consistently outperforms most advanced distillation methods. Our work provides a robust framework for practical knowledge transfer. Future research could extend SRA to embedding model scenarios or other representation learning settings.

\section{Limitations}
Our work was conducted under limited computational budgets, which constrained the scope of experimentation. Moreover, the current logit mapping in our framework is static. While this design proved effective, it inevitably overlooks other important dimensions of the logit space that may carry valuable knowledge. Furthermore, aligning the span representation spaces necessitates on-the-fly teacher inference, since precomputing all teacher span embeddings would require enormous storage. Finally, our experiments were conducted under fixed computational budgets and benchmark conditions. We view these limitations as opportunities for further development, particularly in designing more efficient span-level alignment mechanisms and adaptive mapping strategies that can make SRA both more scalable and more reliable.

% \section*{Acknowledgements}

\section*{Acknowledgments}
Trung Le was supported by the Air Force Office of Scientific Research under award number FA2386-25-1-4023 and the ARC Discovery Project grant DP250100262.

\bibliography{ref}
\bibliographystyle{acl_natbib}

\newpage
\appendix
\newpage
\appendix

\section{Center-of-Mass Dynamics and Transformer Analogy}
\label{sec:CoM}

\begin{table*}[h!]
\centering

\renewcommand{\arraystretch}{1.35}
\setlength{\tabcolsep}{1pt}
\begin{tabular}{@{}l c l@{}}
\toprule
\textbf{Transformer Concept} & $\leftrightarrow$ & \textbf{Multiparticle Dynamical Analog} \\
\midrule
Token Hidden State $h_{\ell,i}$ 
& $\leftrightarrow$ 
& Particle Position $x_{i,t} \in \mathbb{R}^d$ \\[8pt]

Layer Index $\ell$ 
& $\leftrightarrow$ 
& Discretized Time Step $t$ \\[8pt]

Multi-Head Self-Attention 
& $\leftrightarrow$ 
& \textbf{Diffusion:} inter-particle interaction \\
 $\text{MHA}_{W_{\text{att}}^\ell}(h_{\ell,i}, [h_{\ell,1},\dots,h_{\ell,n}])$
 &  &
 $F\!\big(x_{i,t},[x_{1,t}\dots,x_{n,t}]\big) $ \\[8pt]
Feed-Forward Network (per token) 
& $\leftrightarrow$ 
& \textbf{Convection:} private particle update \\
 $\text{FFN}_{W_{\text{ffn}}^\ell}(\tilde{h}_{\ell,i})$
 &  &
 $G(\tilde{x}_{i,t}) $\\[8pt]

Transformer Residual Update
& $\leftrightarrow$ 
& Lie–Trotter splitting step \\
 $\tilde{h}_{\ell,i}
  = h_{\ell,i}
  + \mathrm{MHA}_{W_{\mathrm{att}}^{\ell}}
    \!\big(h_{\ell,i},[h_{\ell,1},\ldots,h_{\ell,n}]\big) $
 &  &
  $\tilde{x}_i = x_i + \gamma F(x_{i,t}[x_{1,t}\ldots,x_{n,t}])$\\
% x_i^{+} = \tilde{x}_i + \gamma G (\tilde{x}_i,t)\\
     $h_{\ell+1,i}
  = \tilde{h}_{\ell,i}
  + \mathrm{FFN}_{W_{\mathrm{ffn}}^{\ell}}
    \!\big(\tilde{h}_{\ell,i}\big)$
    &  &
        $x_i^{+} = \tilde{x}_i + \gamma\,G\!\big(\tilde{x}_{i,t}\big)$\\

\bottomrule
\end{tabular}
\caption{Dictionary aligning Transformer operations with their Multi-Particle Dynamical System (MPDS) counterparts \citep{lu2019understanding}.}
\label{tab:mpds_transformer_equiv}
\end{table*}

Multi-Particle Dynamic Systems (MPDS) model the motion of particle collections in space using differential equations \cite{moulton1970introduction}. Each particle’s behavior is influenced by convection, governing its intrinsic dynamics, and diffusion, capturing interactions with other particles.The Transformer architecture can be interpreted through the lens of a multi-particle dynamic system \cite{lu2019understanding}.

In physics, consider a system of $N$ particles, where each particle has mass $m_i$ and position $x_i$. The center of mass of the system is given by:

\begin{equation*}
    \tilde{x}(t) = \frac{1}{M} \sum_{i = 1}^{N} m_i(t) x_i(t), \quad M = \sum_{i = 1}^{N} m_i(t)
\label{eq:CoM}
\end{equation*}

These expressions extend naturally to the case where the $N$ particles are partitioned into $K$ clusters:
\begin{equation}
\begin{aligned}
    \tilde{x}(t) &= \frac{1}{M} \sum_{k=1}^{K} M_k(t)\, \tilde{x}_k(t), \\
    \tilde{x}_k(t) &= \frac{1}{M_k(t)} \sum_{i \in C_k} m_i(t)\,x_i(t), \\
    &M_k(t) = \sum_{i \in C_k} m_i(t).
\end{aligned}
\label{eq:cluster_CoM}
\end{equation}

Thus, both the center-of-mass can be computed hierarchically through sub-clusters.

\paragraph{Analogy to Transformers.}  
We can establish an analogy between particle dynamics and the evolution of hidden states across Transformer layers. 
Let $l$ denote the layer index, and identify $x_i(t)$ with hidden states $H_{l,i}$. The center-of-mass position of span $k$ at layer $l$ is given by:
\begin{equation}
\begin{aligned}
    (H_l)^{\text{Span}}_k &= \frac{1}{M_{l,k}}\sum_{i \in \text{Span}_k} w_{l,i}\, H_{l,i}, \\
    M_{l,k} &= \sum_{i \in \text{Span}_k} w_{l,i}.
\end{aligned}
\end{equation}
where $(H_l)^{\text{Span}}k$ denotes the span representation, and $w{l,i}$ plays the role of the “mass” assigned to token $i$ at layer $l$.

Hence, by defining span representations as center-of-mass positions, their statescan be computed hierarchically—analogous to clusters of particles. 
This perspective enables us to exploit well-established physical principles to design a systematic and physically grounded mechanism for knowledge transfer.

\section{Experimental Details}
\label{sec: appendix_B}

\paragraph{Training and Evaluation} For GPT2-120M and GPT2-340M, we employ full fine-tuning. For TinyLLaMA, GPT2-1.5B, and OPT2.7B, we adopt LoRA-based fine-tuning. Detailed training configurations for each model are summarized in Table~\ref{tab:training-config}. 
All knowledge distillation (KD) experiments are performed on the Databricks-Dolly-15k dataset. We evaluate across multiple datasets covering different domains and tasks, using validation ROUGE-L \cite{lin2004rouge} for model selection.

\begin{table}[htbp]
\centering
\scriptsize
\renewcommand{\arraystretch}{1.15}
\setlength{\tabcolsep}{2pt}
\begin{tabular}{l|ccccc}
\toprule
\textbf{Student} & \textbf{Epoch} & \textbf{LR} & \textbf{Tuning} & \textbf{LoRA Dropout} & \textbf{LoRA Rank/Alpha} \\
\midrule
\multicolumn{6}{l}{\textit{Teacher: Qwen1.5-1.8B}} \\
\midrule
GPT2-120M  & 20 & $5\text{e-}4$ & Full & -- & -- \\
GPT2-340M  & 20 & $5\text{e-}4$ & Full & -- & -- \\
\midrule
\multicolumn{6}{l}{\textit{Teacher: Mistral-7B}} \\
\midrule
TinyLLaMA  & 15 & $1\text{e-}3$ & LoRA & 0.1 & 16/64 \\
\midrule
\multicolumn{6}{l}{\textit{Teacher: Qwen2.5-7B-Instruct}} \\
\midrule
GPT2-1.5B  & 15 & $1\text{e-}3$ & LoRA & 0.1 & 16/64 \\
OPT-2.7B   & 15 & $1\text{e-}3$ & LoRA & 0.1 & 16/64 \\
\bottomrule
\end{tabular}
\caption{Training configurations. Projector LR = $5\text{e-}4$, 
batch size = 8, and cosine LR scheduler are shared across all settings.}
\label{tab:training-config}
\end{table}
\paragraph{Hyperparameter}
The hyperparameters are reported in Table~\ref{tab:best-alpha}. 
Throughout all experiments, we set the KL-divergence temperature $\tau$ to $2.0$, hyperparameter $p$ to $1.0$ and the weight $\lambda$ of the geometric regularization constraint to $50$.

\begin{table}[hb]
\centering
\scriptsize
\renewcommand{\arraystretch}{1.15}
\setlength{\tabcolsep}{2pt}
\begin{tabular}{|c|c|c|c|c|}
\hline
\textbf{GPT2-120M} & \textbf{GPT2-340M} & \textbf{TinyLLaMA} & \textbf{GPT2-1.5B} & \textbf{OPT2.7B} \\
\hline
 0.5 & 0.5 & 0.6 & 0.6 & 0.8 \\
\hline
\end{tabular}%
\caption{The hyperparameters $\alpha$ for different configurations}
\label{tab:best-alpha}
\end{table}

\section{Additional Ablation Results}
\label{sec: Appendix_C}
\begin{table}[ht]
\centering
\scriptsize
\renewcommand{\arraystretch}{1.15}
\setlength{\tabcolsep}{2pt}
\begin{tabular}{l|cccccc}
\toprule
\multirow{0}{*}{\textbf{Hyperparameter $p$}}
& \multicolumn{6}{c}{Teacher: Qwen1.5-1.8B $\rightarrow$ GPT-2 120M}\\
\cmidrule(lr){2-7}
& \textbf{Dolly} & \textbf{Vicuna} & \textbf{Self Inst} & \textbf{S-NI} & \textbf{Dialog} & \textbf{Avg.} \\
\midrule
$p = 0.0$       &  21.54 & 13.49 & 11.09 & 20.69 & 12.62 & 15.89 \\
$p = 0.5$       & \textbf{23.75}  & 15.52 & 12.92 & \textbf{25.08} & 12.66 & \textbf{17.99}\\
$p = 1.0$       & 23.36 & \textbf{16.08} & \textbf{13.16} & 23.70 & \textbf{13.56} & 17.97  \\
\bottomrule
\end{tabular}
\caption{Effect of the sharpness hyperparameter $p$ on model performance.}
\label{tab:p_param}
\end{table}

\subsection{Effect of Hyperparameter $p$}
The results in ~\ref{tab:p_param} demonstrate that applying weighting through the sharpness parameter $p$ consistently improves model performance, with the best average achieved at $p=0.5$. This suggests that moderate weighting helps the student model better align with the teacher’s distribution. However, further increasing $p$ to $1.0$ brings no improvement, suggesting a weighting threshold beyond which the model no longer benefits from sharper distributions.

\subsection{Effect of Transferred Layers} 
Increasing the number of transferred layers consistently reduces performance (Table~\ref{tab:n_transferred}). 
We hypothesize that transferring from too many intermediate layers amplifies architectural and representational
mismatches between the teacher (Qwen1.5) and the student (GPT-2), 
leading to unstable supervision and degraded knowledge transfer efficiency.

\begin{table}[ht]
\centering
\scriptsize
\renewcommand{\arraystretch}{1.15}
\setlength{\tabcolsep}{2pt}
\begin{tabular}{l|cccccc}
\toprule
\multirow{0}{*}{\textbf{Transferred Layers}}
& \multicolumn{6}{c}{Teacher: Qwen1.5-1.8B $\rightarrow$ GPT-2 120M}\\
\cmidrule(lr){2-7}
& \textbf{Dolly} & \textbf{Vicuna} & \textbf{Self Inst} & \textbf{S-NI} & \textbf{Dialog} & \textbf{Avg.} \\
\midrule
0      & \textbf{23.74} & 15.85 & 12.03 & 22.11 & 11.78 & 17.10 \\
1 (12)      & 23.36 & \textbf{16.08} & \textbf{13.16} & \textbf{23.70} & \textbf{13.56} & \textbf{17.97} \\
3 (10, 11, 12)      & 21.13 & 13.61 & 11.61 & 20.28 & 10.64 & 15.45 \\
5 (8, 9, 10, 11, 12) & 20.78 & 12.27 & 10.81 & 20.47 & 11.67 & 15.20 \\
\bottomrule
\end{tabular}
\caption{Effect of the number of transferred layers on model performance.}
\label{tab:n_transferred}
\end{table}

% \begin{figure}[h!]
%     \centering
%     \includegraphics[width=1\linewidth]{gpt4.1 eval.png}
%     \caption{Win rates (\%) for distilling Mistral 7B to TinyLlama 1.1B , evaluated by GPT-4.1 on response quality.
%     }
%     \label{fig:placeholder}
% \end{figure}
\subsection{The effect of Span-Level Hidden State Transfer without Span-Level Logits loss}
\label{sec:sra_wo_logit}
\begin{table}[h!]
\centering
\footnotesize
\renewcommand{\arraystretch}{1.15}
\setlength{\tabcolsep}{1.5pt}
\begin{tabular}{l|cccccc}
\toprule
\textbf{Methods} & \textbf{Dolly} & \textbf{Vicuna} & \textbf{SelfInst} & \textbf{S-NI} & \textbf{Dialog} & \textbf{Avg.}\\
\specialrule{1.0pt}{1.0pt}{1.0pt}
\multicolumn{7}{c}{\textit{Qwen1.5--1.8B} $\rightarrow$ \textit{GPT-2 120M}}\\
\midrule
Teacher                          & 28.23 & 19.59 & 19.58 & 34.36 & 14.18 & 23.19 \\
\midrule
SFT                              & 23.78 & 17.04 &  6.78 &  7.81 &  8.29 & 12.74 \\
ULD                              & 23.77 & 14.33 &  9.30 & 14.04 &  8.63 & 14.01 \\
MinED                            & 24.21 & 14.96 & 10.02 & 16.40 &  9.79 & 15.08 \\
MultiLevelOT                     & 23.02 & 13.79 &  8.41 & 12.26 &  8.79 & 13.25 \\
DSKD                             & \textbf{24.26} & 15.25 & 10.07 & 17.15 & 10.03 & 15.35 \\
SRA w/o $\mathcal{L}_{KD}^{Span}$ & 23.24 & 15.18 & 11.02 & 21.45 &  9.32 & 16.04 \\
\rowcolor{gray!20}
\textbf{SRA}                     & 23.36 & \textbf{16.08} & \textbf{13.16} & \textbf{23.70} & \textbf{13.56} & \textbf{17.97} \\
\specialrule{1.0pt}{1.0pt}{1.0pt}
\multicolumn{7}{c}{\textit{Qwen1.5--1.8B} $\rightarrow$ \textit{GPT-2 340M}}\\
\midrule
Teacher                          & 28.23 & 19.59 & 19.58 & 34.36 & 14.18 & 23.19 \\
\midrule
SFT                              & 23.11 & 14.89 &  9.09 & 13.03 &  8.00 & 13.62 \\
ULD                              & 23.90 & 15.04 &  9.96 & 16.26 &  8.76 & 14.78 \\
MinED                            & 24.48 & 15.56 & 11.21 & 15.69 &  8.98 & 15.18 \\
MultiLevelOT                     & 23.95 & 14.80 & 10.21 & 15.87 &  8.99 & 14.76 \\
DSKD                             & 25.43 & 15.08 & 11.29 & 17.18 &  8.90 & 15.57 \\
SRA w/o $\mathcal{L}_{KD}^{Span}$ & \textbf{25.56} & 15.93 & 13.28 & 23.70 &  8.95 & 17.48 \\
\rowcolor{gray!20}
\textbf{SRA}                     & 23.97 & \textbf{16.31} & \textbf{13.64} & \textbf{24.49} & \textbf{12.08} & \textbf{18.10} \\
\bottomrule
\end{tabular}
\caption{ROUGE-L comparison of SRA and SRA w/o $\mathcal{L}_{KD}^{Span}$ against CTKD baselines.}
\label{tab:sra_wo_logit}
\end{table}

Table~\ref{tab:sra_wo_logit} reports the performance of SRA with and 
without the span-level logit distillation loss $\mathcal{L}_{KD}^{Span}$ 
across two teacher--student pairs. Several observations are noteworthy. 
First, even without $\mathcal{L}_{KD}^{Span}$, SRA already surpasses all 
CTKD baselines in average ROUGE-L (16.04 vs.\ 15.35 for GPT-2 120M; 
17.48 vs.\ 15.57 for GPT-2 340M), demonstrating that the span-level 
hidden-state alignment and geometric regularization alone constitute a 
strong and tokenizer-agnostic distillation signal. Second, incorporating 
$\mathcal{L}_{KD}^{Span}$ consistently yields further gains across nearly 
all benchmarks, with the most pronounced improvements on Dialog 
(+4.24 for GPT-2 120M; +3.13 for GPT-2 340M) and SelfInst, confirming 
that logit distillation over the shared vocabulary provides a 
complementary supervisory signal that reinforces the teacher's predictive 
behavior. These results collectively validate the design choice of 
combining tokenizer-agnostic span representations with vocabulary-intersection 
logit distillation in the full SRA objective.
\subsection{Comparison with ALM}
\label{app:alm}

\begin{table}[h!]
\centering
\footnotesize
\renewcommand{\arraystretch}{1.15}
\setlength{\tabcolsep}{1.5pt}
\begin{tabular}{l|cccccc}
\toprule
\textbf{Methods} & \textbf{Dolly} & \textbf{Vicuna} & \textbf{SelfInst} & \textbf{S-NI} & \textbf{Dialog} & \textbf{Avg.}\\
\specialrule{1.0pt}{1.0pt}{1.0pt}
\multicolumn{7}{c}{\textit{Qwen1.5--1.8B} $\rightarrow$ \textit{GPT-2 120M}}\\
\midrule
Teacher & 28.23 & 19.59 & 19.58 & 34.36 & 14.18 & 23.19 \\
\midrule
SFT     & 23.78 & 17.04 &  6.78 &  7.81 &  8.29 & 12.74 \\
ALM     & 20.86 & 14.76 & 10.65 & 17.76 & 10.44 & 14.89 \\
\rowcolor{gray!20}
\textbf{SRA} & \textbf{23.36} & \textbf{16.08} & \textbf{13.16} & \textbf{23.70} & \textbf{13.56} & \textbf{17.97} \\
\specialrule{1.0pt}{1.0pt}{1.0pt}
\multicolumn{7}{c}{\textit{Qwen2.5-7B-Instruct} $\rightarrow$ \textit{GPT2-1.5B}}\\
\midrule
Teacher & 28.49 & 20.48 & 24.67 & 39.87 & 16.86 & 26.07 \\
\midrule
SFT     & 21.83 & 15.95 & 13.62 & 21.66 & 10.91 & 16.79 \\
ALM     & 25.78 & 16.36 & 15.63 & 25.78 & 11.07 & 18.92 \\
\rowcolor{gray!20}
\textbf{SRA} & \textbf{26.45} & \textbf{18.25} & \textbf{17.22} & \textbf{29.50} & \textbf{13.52} & \textbf{20.99} \\
\bottomrule
\end{tabular}
\caption{ROUGE-L comparison of SRA against ALM across representative 
teacher–student pairs. SRA outperforms ALM in all settings.}
\label{tab:alm_comparison}
\end{table}
As shown in Table~\ref{tab:alm_comparison}, \textbf{SRA} consistently 
outperforms ALM across all benchmarks and both teacher–student configurations. 
While ALM aggregates over spans at the token-likelihood level, it still 
relies on token-level predictive traces that are sensitive to tokenizer 
discrepancies. In contrast, SRA operates directly on span-level hidden 
states with geometric regularization, bypassing token-level misalignment 
entirely. This design difference is reflected in the results: SRA achieves 
an average ROUGE-L gain of \textbf{+3.08} over ALM on the 
Qwen1.5--1.8B $\rightarrow$ GPT-2 120M pair and \textbf{+2.07} on the 
Qwen2.5-7B-Instruct $\rightarrow$ GPT2-1.5B pair, with pronounced improvements on benchmarks such as S-NI and Dialog, suggesting that span-level hidden-state alignment generalizes more robustly across instruction-following tasks.

\section{Longest Common Subsequence-based Span Alignment}
\label{sec: Appendix_D}

\begin{algorithm}[ht]
\caption{LCS-based Span Alignment}
\label{alg:span_alignment}
\KwIn{\\ 
\hspace*{2.5em} Teacher offsets $O^T$, \\
\hspace*{2.5em} Student offsets $O^S$, \\
\hspace*{2.5em} First non-special token indices $t^T_0$, $t^S_0$}

\KwOut{List of matched span indices $Ms$}\vspace{0.5em}

$m \gets \text{len}(O^T)$, $n \gets \text{len}(O^S)$\;

$i \gets 0$, $j \gets 0$; 

$aligns \gets [(t^T_0, t^S_0)]$, $Ms \gets [\,]$\;

\While{$i < m$ \textbf{and} $j < n$}{
    \If{$O^T[i] = 0$}{
        $i \gets i+1$\;
        \textbf{continue}\;
    }
    \If{$O^S[j] = 0$}{
        $j \gets j+1$\;
        \textbf{continue}\;
    }
    \eIf{$O^T[i] = O^S[j]$}{
        $span^S = (aligns[-1][0], i)$\;
        $span^T = (aligns[-1][1], j)$\;
        append $(span^S, span^T)$ to $Ms$\;
        append $(i+1, j+1)$ to $aligns$\;
        $i \gets i+1$,\, $j \gets j+1$\;
    }{
        \eIf{$O^T[i] < O^S[j]$}{
            $i \gets i+1$\;
        }{
            $j \gets j+1$\;
        }
    }
}
\Return{$Ms$}\;
\end{algorithm}

\vspace{60mm}
\section{Prompt for evaluation via GPT-4}

\begin{figure}[htbp]
\centering
\begin{tcolorbox}[colback=promptbg, colframe=black, boxrule=0.5pt,
                  left=3pt,right=3pt,top=3pt,bottom=3pt,
                  width=1\columnwidth] % <= shrink width here
\scriptsize % <= smaller text; try \footnotesize if you prefer
\raggedright
\textbf{Please act as an impartial judge} and compare the quality of response A and response B provided by two AI assistants to the user question displayed below.
Focus on how natural, fluent, and human-like the language sounds
Your evaluation should prioritize effectiveness, clarity, readability, technical accuracy and completeness.

\medskip
  - If A is significantly better, answer "A".\\
  - If B is significantly better, answer "B".\\
  - If both are similar in quality (both bad or both good or show no significant difference), answer "Tied".\\

\medskip
\textbf{[Question]}\\
\{question or instruction\}

\medskip
\textbf{[Response A]}\\
\{response A\}

\medskip
\textbf{[Response B]}\\
\{response B\}
\end{tcolorbox}
\caption{Prompt for GPT-4 evaluation.}
\label{fig:gpt4}
\vspace{-2mm}
\end{figure}

\cleardoublepage

\end{document}